\pdfoutput=1
\documentclass[11pt]{article}

\usepackage[preprint]{acl}

\usepackage{times}
\usepackage{latexsym}

\usepackage[T1]{fontenc}
\usepackage[utf8]{inputenc}

\usepackage{microtype}

\newcommand\eztb[1]{\begin{tabular}[t]{@{}l@{}}#1\end{tabular}}

\usepackage{tabularray}

\usepackage{inconsolata}

\usepackage{graphicx}

\title{TookaBERT: A Step Forward for Persian NLU}

\author{
 \textbf{MohammadAli SadraeiJavaheri},
 \textbf{Ali Moghaddaszadeh},
 \textbf{Milad Molazadeh},
 \textbf{Fariba Naeiji},
 \\
 \textbf{Farnaz Aghababaloo},
 \textbf{Hamideh Rafiee},
 \textbf{Zahra Amirmahani},
 \textbf{Tohid Abedini},
 \\
 \textbf{Fatemeh Zahra Sheikhi},
 \textbf{Amirmohammad Salehoof}.
 \\
 Part Artificial Intelligence Research Center
}

\begin{document}
\maketitle
\begin{abstract}
The field of natural language processing (NLP) has seen remarkable advancements, thanks to the power of deep learning and foundation models. Language models, and specifically BERT, have been key players in this progress. In this study, we trained and introduced two new BERT models using Persian data. We put our models to the test, comparing them to seven existing models across 14 diverse Persian natural language understanding (NLU) tasks. The results speak for themselves: our larger model outperforms the competition, showing an average improvement of at least $+2.8$ points. This highlights the effectiveness and potential of our new BERT models for Persian NLU tasks.
\end{abstract}

\section{Introduction}

Recent advancements in natural language processing (NLP) are largely attributed to deep learning methods and the evolution of foundation models. Language models have played a pivotal role in this progress, with a standard approach of first training a language model and then fine-tuning it for specific NLP tasks\citep{howard-ruder-2018-universal, peters_deep_2018}. This method has become very popular in the field. Specifically after The transformer architecture, introduced by \citet{vaswani_attention_2017}, revolutionized these models. BERT, developed by \citet{devlin_bert_2019}, and GPT, created by \citet{Radford2018ImprovingLU}, were the first successful models to combine the transformer architecture with pre-training as a language model. A key difference between these two models is their approach to attention; GPT utilizes causal attention, while BERT employs a bidirectional approach. This distinction makes the GPT model more suitable for generative tasks.

Generative tasks quickly embraced GPT as the model of choice, while understanding tasks found greater success with BERT. The release of GPT-3 \citep{NEURIPS2020_1457c0d6} further boosted the popularity of generative models in natural language processing (NLP). With the introduction of ChatGPT, generative models have taken center stage. These models require a large number of parameters to perform well across different tasks. While they offer impressive capabilities and versatility, they often necessitate substantial computational power and resources. Training or fine-tuning such large models on standard computers is not feasible.

In contrast, smaller encoder models have demonstrated remarkable performance in natural language understanding tasks. These models can be fine-tuned and executed with far less hardware, and their performance after fine-tuning is comparable to that of a large language model on specific tasks.

In this study, we address the lack of a large-scale Persian BERT model by training a new model using state-of-the-art techniques on a large Persian dataset. We also trained a base version to serve as a comparison point. Our main contributions are twofold: \textbf{(I)} We are making available to the public two new checkpoints of the BERT model specifically trained for the Persian language, which fills a significant gap in the availability of large-scale models for this language; \textbf{(II)} Through systematic evaluation across 14 diverse Persian natural language understanding (NLU) tasks, our larger checkpoint demonstrates superior performance, achieving an average improvement of at least $+2.8$ points compared to other models.

\section{Related Works}

\begin{table*}[t]

\begin{center}
\resizebox{\textwidth}{!}{

\begin{tblr}{
    column{2-9} = {c},
    vline{2, 7} = {1-3}{},
    vline{9} = {-}{},
    cell{2}{2,3,4} = {}{b},
    cell{1}{2} = {c=5}{c},
    cell{1}{7} = {c=2}{c},
    hline{2, 11} = {-}{},
    hline{4, 13} = {-}{2px},
}
  & ParsiNLU &&&&& NER \\
Task→ & \eztb{Reading\\Comprehension} & \eztb{Sentiment\\Analysis} & QQP & \eztb{Multiple\\Choice} & \eztb{Entailment} & \eztb{Peyma NER} & \eztb{MultiConer} & Avg.\\
Models↓  Metrics→ & EM/F1 & F1/Acc. & F1/Acc. & F1/Acc. & F1/Acc. & F1/Acc. & F1/Acc. & -\\
XLM-V & 8.0/26.7 & 82.3/84.1 & 80.1/80.0 & 35.3/35.2 & 28.3/37.8 & \textbf{88.1/97.8} & 58.8/92.2 & 59.6\\
XLM-RoBERTa-Base & 20.0/40.4 & 89.7/89.9 & 79.1/79.0 & 32.4/32.4 & 53.8/53.7 & 87.8/97.8 & 60.4/92.5 & 64.9\\
Bert Multilingual & 27.2/42.2 & 86.8/87.5 & 79.2/79.3 & 33.4/33.1 & 54.4/54.4 & 83.7/96.9 & 60.3/92.5 & 65.1\\
Shiraz & 17.6/39.6 & 87.1/87.8 & 79.7/79.5 & 34.7/34.5 & 39.9/41.5 & 86.4/97.6 & 59.1/92.8 & 62.7\\
Parsbert & 20.0/39.6 & 88.6/88.9 & 80.2/80.1 & 35.3/35.2 & 49.9/49.6 & 86.8/97.7 & 64.9/93.2 & 65.0\\
AriaBERT & 14.4/35.5 & 87.9/88.4 & 79.1/78.8 & 30.8/30.9 & 44.5/44.4 & 84.1/97.3 & 61.0/92.5 & 62.1\\
FaBERT & 27.2/48.4 & 90.4/90.5 & 82.3/82.3 & 32.5/32.4 & \textbf{55.0/54.8} & 87.4/97.6 & 63.9/93.0 & 67.0\\
TookaBERT-Base & 20.8/42.5 & 89.6/89.7 & 81.3/81.3 & 33.6/33.8 & 49.3/48.9 & 86.3/97.8 & 66.2/93.3 & 65.3\\
TookaBERT-Large & \textbf{33.6/60.5} & \textbf{91.1/91.4} & \textbf{82.7/82.6} & \textbf{36.1/36.0} & 54.3/54.1 & 86.2/98.0 & \textbf{69.7/94.1} & \textbf{69.3}\\

\end{tblr}

}
\end{center}

\caption{\label{table_r1}
Performance comparison of various base models on different end-tasks. The 'Average' column is calculated based on the scores presented in this table for both ParsiNLU and named entity recognition tasks.}
\end{table*}

BERT was introduced by \citep{devlin_bert_2019} and improved by RoBERTa \citep{liu_roberta_2019}. The key difference between these models is their training objective. RoBERTa removed the next sentence prediction loss, focusing solely on masked language modeling loss, and demonstrated that this alone is sufficient for training.

There are several pre-trained Persian BERT models, with the earliest and most well-known being ParsBERT by \citet{parsbert_model}. Other models include AriaBERT\citep{ariabert_model}, FaBERT\citep{masumi_fabert_2024}, and the recent Shiraz and Tehran models by \citet{shiraz_tehran}. While Persian BERT models from last year outperformed ParsBERT, none of them experimented with model sizes beyond BERT-Base, such as BERT-Large. Multilingual pre-trained BERT models have also exhibited good performance on Persian tasks \citep{conneau_unsupervised_2020, liang_xlm-v_2023}.

\section{Data}

\noindent\textbf{Pre-training dataset:} This includes a mix of public datasets: hmblogs \citep{khansari2021hmblogs}, the cleaned Persian subset of madlad \citep{kudugunta2023madlad400}, and PersianWebScrapper \citep{targoman_pws}. We normalized these datasets using NFKC to ensure standard UTF-8 encoding for Persian text. We also replaced some Arabic characters with their Persian equivalents.

\noindent\textbf{Evaluation dataset:} We aggregated multiple Persian NLU datasets for evaluation. Unlike English, which has well-known collections like GLUE \citep{glue_ds} and SuperGLUE \citep{superglue_ds}, Persian lacks a comprehensive NLU dataset collection. We used the ParsiNLU dataset \citep{parsinlu_ds}, which includes 6 tasks, one of which is machine translation and is not relevant for BERT evaluation. Therefore, we utilized the remaining 5 tasks from ParsiNLU along with the following individual datasets: PeymaNER \citep{peyma_ds}, Persian subset of MultiCoNER \citep{multiconer_ds}, PQA \citep{pqa_ds}, PQuAD \citep{pquad_ds}, FarsTail \citep{farstail_ds}, DeepSentiPers \citep{dsp_ds}, SnappFood sentiment dataset\footnote{Available at: \url{https://bit.ly/2Xu2xq1}} \citep{parsbert_model}, SentiPers \citep{sp_ds}, and Arman Emotion \citep{arman_emotion}. Our aim is to give a fair comparison for the foundation models we are testing by using a variety of Persian language tasks.

\section{Methodology}

\begin{table*}[t]

\begin{center}
\resizebox{\textwidth}{!}{

\begin{tblr}{
    column{2-9} = {c},
    cell{1}{2} = {c=2}{c},
    cell{1}{4} = {}{c},
    cell{1}{5} = {c=4}{c},
    vline{2, 4, 5} = {1-3}{},
    vline{9} = {-}{},
    hline{2} = {2-9}{},
    hline{2, 11} = {-}{},
    hline{4, 13} = {-}{2px},
}
  & QA &&NLI & Sentiment &&\\
Task→ & \eztb{PQA} & \eztb{PQuAD} & \eztb{FarsTail} & \eztb{DeepSentiPers} & \eztb{Snappfood} & \eztb{SentiPers} & \eztb{Arman\\Emotion} & Avg.\\
Models↓  Metrics→ & EM/F1/Has-EM/Has-F1 & EM/F1/Has-EM/Has-F1 & F1/Acc. & F1/Acc. & F1/Acc. & F1/Acc. & F1/Acc. & -\\
XLM-V & 40.3/57.0/29.8/62.3 & 73.3/85.7/68.2/85.6 & 81.1/81.2 & 83.4/83.4 & 88.0/88.1 & 74.6/74.8 & 42.3/48.0 & 71.1\\
XLM-RoBERTa-Base & 39.2/58.0/30.2/62.5 & 73.7/86.2/68.2/85.8 & 82.0/82.0 & 84.0/84.1 & 87.6/87.7 & 75.7/75.8 & 69.8/70.0 & 75.0\\
Bert Multilingual & 37.8/56.0/28.1/60.9 & 71.8/84.7/65.9/84.0 & 82.7/82.8 & 78.6/78.7 & 87.1/87.1 & 71.7/71.9 & 62.2/62.4 & 72.1\\
Shiraz & 30.4/44.4/18.1/51.6 & 66.0/81.2/59.6/81.3 & 77.8/77.8 & 81.2/81.1 & 87.1/87.1 & 73.4/73.8 & 71.6/71.7 & 71.3\\
Parsbert & 36.7/49.8/26.8/56.1 & 71.4/84.2/66.3/84.6 & 80.9/80.9 & 80.2/80.2 & 87.6/87.7 & 74.3/74.4 & 67.2/67.4 & 72.8\\
AriaBERT & 35.0/44.4/19.5/48.1 & 68.1/81.2/62.1/80.9 & 74.5/74.4 & 80.5/80.5 & 87.7/87.8 & 73.6/73.8 & 71.4/71.5 & 71.1\\
FaBERT & 39.8/55.6/30.0/62.2 & 72.6/85.4/67.2/85.3 & 83.7/83.7 & 82.7/82.7 & \textbf{88.1/88.1} & 75.0/75.3 & 72.9/73.0 & 75.3\\
TookaBERT-Base & 38.7/54.4/29.2/60.9 & 73.2/85.7/68.3/85.9 & 83.3/83.4 & 83.9/83.9 & 87.4/87.4 & 74.9/75.1 & 73.2/73.4 & 75.3\\
TookaBERT-Large & \textbf{46.2/65.0/38.4/69.9} & \textbf{75.6/88.1/70.2/87.8} & \textbf{89.7/89.7} & \textbf{85.7/85.8} & \textbf{88.1/88.1} & \textbf{76.3/76.5} & \textbf{74.7/74.7} & \textbf{78.6}\\

\end{tblr}

}
\end{center}

\caption{\label{table_r2}
Performance comparison of base models on various end-tasks. The 'Average' column is calculated based on the scores from question answering, natural language inference, and sentiment analysis datasets, as presented in this table.}
\end{table*}

\subsection{Tokenizer}
The original BERT paper used the WordPiece algorithm for tokenization \citep{devlin_bert_2019}. However, newer models mostly use byte pair encoding (BPE) \citep{bpe_paper}. Examples include RoBERTa \citep{liu_roberta_2019} and LLaMA \citep{llama_paper}. We also chose to use BPE for our tokenizer. We trained a new Persian tokenizer using this algorithm and set our vocab size to 48,000. During the training phase, we also used BPE-dropout \citep{bpe-drop} of 0.1 to regularize our model and improve generalization over unseen tokens.

\subsection{BERT Pre-training}
We trained our BERT model using the latest advancements and ideas in the field. Our approach was influenced by other works such as MosaicBERT \citep{portes2023mosaicbert}. To speed up training and efficiently utilize GPU memory, we employed flash attention v2 \citep{dao_flashattention-2_2023}. Additionally, we made use of multiple GPUs effectively by utilizing the ZeRO stage 2 optimizer \citep{rajbhandari_zero_2020}.

Similar to RoBERTa \citep{liu_roberta_2019}, we used only mask language modeling as our objective and excluded next sentence prediction. To avoid the computational waste of padding tokens, we concatenated our entire dataset and used it as a single batch \citep{unpading_paper}. We also incorporated whole-word-masking \citep{cui_pre-training_2021} to make predicting masks more challenging and improve the model's capabilities.

We trained two BERT models in standard \textbf{Base} and \textbf{Large} sizes. The training was conducted using 8xA100 40GB GPUs.

\subsection{Evaluation}
For a fair comparison, we fine-tuned each model with four different learning rates: 3e-5, 3e-6, 7e-5, and 7e-6 for eight epochs. We then reported the best evaluation score achieved by each model across these learning rates and epochs. In addition to our models, we also evaluated the following pre-trained language models: ParsBERT \citep{parsbert_model}, AriaBERT \citep{ariabert_model}, FaBERT \citep{masumi_fabert_2024}, Shiraz \citep{shiraz_tehran}, BERT multilingual \citep{devlin_bert_2019}, XLM-RoBERTa-Base \citep{conneau_unsupervised_2020}, and XLM-V \citep{liang_xlm-v_2023}

\section{Results}
We employed various evaluation metrics depending on the nature of the task. The score for each task was computed by averaging the respective metrics, with the F1-score and accuracy being the most prevalent across all tasks.

The results, as shown in Tables \ref{table_r1} and  \ref{table_r2}, indicate that TookaBERT-Large demonstrated superior performance compared to other models in the majority of end-tasks. On average, it achieved an impressive 2.8-point improvement over models such as FaBERT.

\section{Conclusion}

In this study, we presented two new BERT models specifically trained for Persian natural language understanding (NLU) tasks. By employing state-of-the-art techniques and a large Persian dataset, we successfully trained and introduced TookaBERT-Base and TookaBERT-Large to address the lack of large-scale Persian BERT models. Our comprehensive evaluations across 14 diverse Persian NLU tasks showcased the superiority of our larger model, outperforming other existing models by an average of $+2.8$ points. This highlights the effectiveness and potential of our new BERT models for Persian language processing.

The significant performance of TookaBERT-Large underscores the importance of model size and training techniques in language model development. By making our models publicly available, we hope to contribute to the advancement of Persian NLU and encourage further research in this direction. Our work showcases the potential of adapting foundation models to specific languages, and we believe that future studies can build upon our models to further enhance their performance and applicability in various Persian language tasks.

\bibliography{custom}

\end{document}